\documentclass[10pt,conference,a4paper]{IEEEtran}
\usepackage[top=0.78in, bottom=1.61in, left=0.53in, right=0.53in]{geometry}
\ifCLASSOPTIONcompsoc
  \usepackage[nocompress]{cite}
\else
  \usepackage{cite}
\fi
\ifCLASSINFOpdf
  \usepackage[pdftex]{graphicx}
\else
\fi
\usepackage{amsmath}
\usepackage{algorithmic}
\usepackage{tablefootnote}
\usepackage{xcolor}
\ifCLASSOPTIONcompsoc
 \usepackage[caption=false,font=footnotesize,labelfont=sf,textfont=sf]{subfig}
\else
 \usepackage[caption=false,font=footnotesize]{subfig}
\fi
\usepackage{url}

\usepackage{multirow}
\usepackage{bm}
\usepackage{algorithm}
\usepackage{mathrsfs}
\usepackage{epstopdf}
\usepackage{threeparttable}
\usepackage{bbding}
\usepackage{tabularx}
\usepackage{booktabs}  

\newcolumntype{I}{!{\vrule width 0.5pt}}
\newlength\savedwidth

\newlength\savewidth

\begin{document}

\title{Adaptive Embedding Gate for Attention-Based \\
Scene Text Recognition}

\author{\IEEEauthorblockN{Xiaoxue Chen$^1$, Tianwei Wang$^1$, Yuanzhi Zhu$^1$, Lianwen Jin$^{12\ast}$, Canjie Luo$^{1}$}
\IEEEauthorblockA{
$^1$College of Electronic and Information Engineering, South China University of Technology, Guangzhou, China\\
$^2$SCUT-Zhuhai Institute of Modern Industrial Innovation, Zhuhai, China \\
xxuechen@foxmail.com, wangtw@foxmail.com, z.yuanzhi@foxmail.com, $^\ast$lianwen.jin@gmail.com, canjie.luo@gmail.com}
}

\maketitle
\begin{abstract}
Scene text recognition has attracted particular research interest because it is a very challenging problem and has various applications. 
The most cutting-edge methods are attentional encoder-decoder frameworks that learn the alignment between the input image and output sequences.
In particular, the decoder recurrently outputs predictions, using the prediction of the previous step as a guidance for every time step.
In this study, we point out that the inappropriate use of previous predictions in existing attention mechanisms restricts the recognition performance and brings instability.
To handle this problem, we propose a novel module, namely adaptive embedding gate (AEG). 
The proposed AEG focuses on introducing high-order character language models to attention mechanism by controlling the information transmission between adjacent characters. 
AEG is a flexible module and can be easily integrated into the state-of-the-art attentional methods.
We evaluate its effectiveness as well as robustness on a number of standard benchmarks, including the IIIT$5$K, SVT, SVT-P, CUTE$80$, and ICDAR datasets.
Experimental results demonstrate that AEG can significantly boost recognition performance and bring better robustness.
\end{abstract}

\IEEEpeerreviewmaketitle

\section{Introduction}
\label{introduction}
In natural scenes, text appears on various kinds of objects, e.g. signboards, road signs and product packagings.
Accurate and rich semantic information carried by the text is important for many application scenarios such as image searching, intelligent inspection, product recognition and autonomous driving. 
For these reasons, scene text recognition has been an active research field in computer vision \cite{guo2016convolutional}, \cite{neumann2012real}, \cite{shi2017end}, \cite{cluo2019moran}, \cite{gao2019reading}.

Although optical character recognition in scanned documents has been considered as a solved problem \cite{nagy2000twenty}, \cite{zhou2014perspective}, recognizing text in natural images is still challenging.
Because the imperfect imagery conditions, such as the aspects of illumination, low resolution and motion blurring limit computers from accurately reading text in the wild.
Furthermore, the various fonts and distorted patterns of irregular text can cause additional challenges in recognition.

\begin{figure}[t]
\centering
\includegraphics[width=0.5\textwidth]{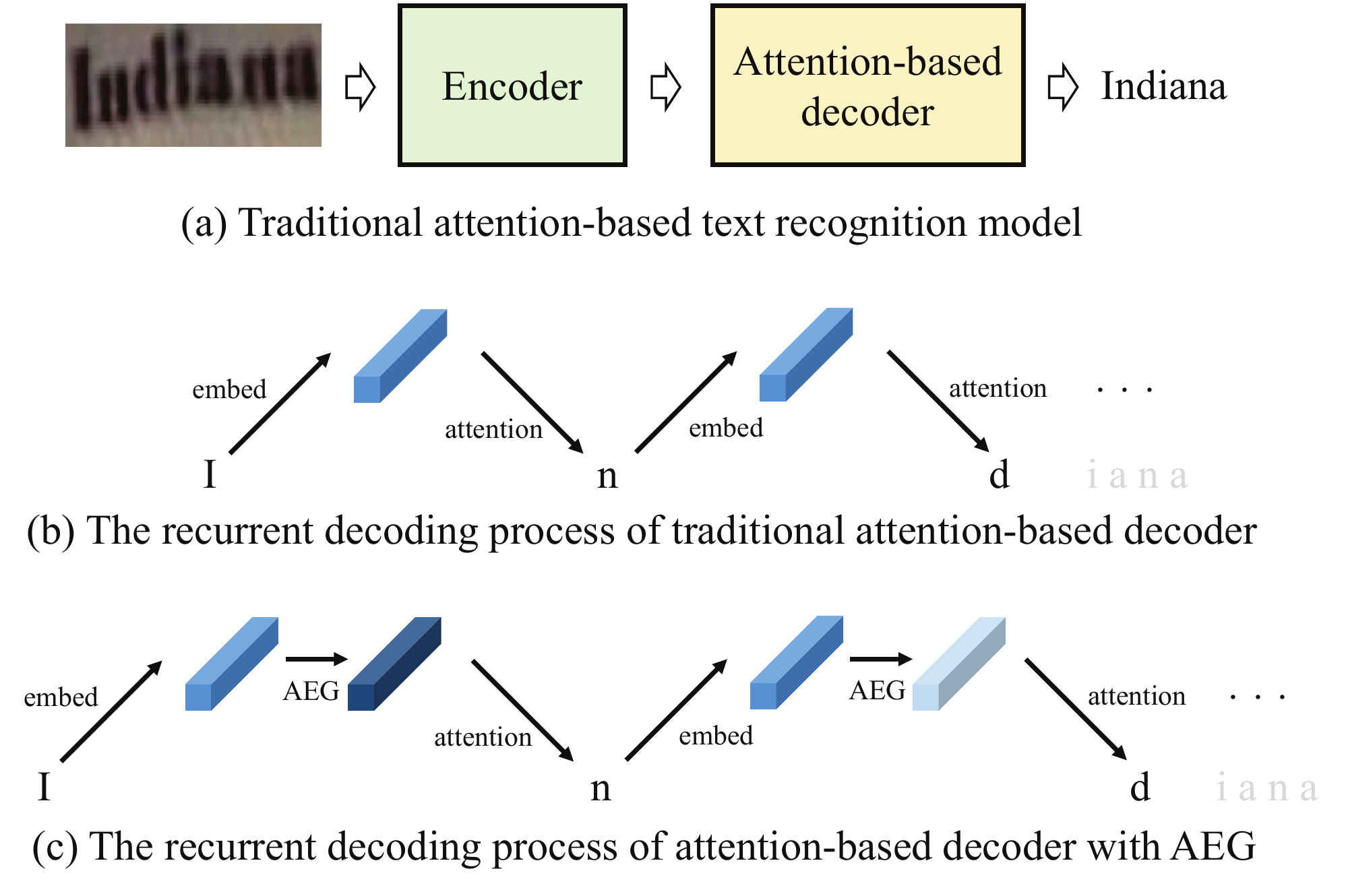}
\caption{(a): Schematic overview of traditional attention-based text recognition model.
(b) and (c): Comparison of the recurrent decoding process between attention-based decoder without/with the proposed AEG.
}
\label{Figure_Compare}
\end{figure}

In recent years, benefiting from the development in deep learning, a large number of scene text recognition methods \cite{wang2011end}, \cite{bissacco2013photoocr}, \cite{shi2017end}, \cite{lee2016recursive}, \cite{shi2018aster} have been reported in the literature with notable success.
As shown in Figure~\ref{Figure_Compare} (a), the famous encoder-decoder frameworks are widely adopted to translate a visual image into a string sequence.
Generally, in the encoding stage, the convolutional neural networks (CNN) are used to extract features from the input image, whereas in the decoding stage, the encoded feature vectors are transcribed into target strings by exploiting the  recurrent neural network (RNN) \cite{cho2014properties}, \cite{hochreiter1997long}, connectionist temporal classification (CTC) \cite{graves2006connectionist} or attention mechanism \cite{bahdanau2014neural}.
In particular, the attention-based approaches \cite{cluo2019moran}, \cite{shi2018aster}, \cite{shi2016robust}, \cite{cheng2017focusing}, \cite{bai2018edit} often achieve better performance owing to the focus on informative areas.

\subsection{Motivation}
As illustrated in Figure~\ref{Figure_Compare} (a) and (b), in the traditional attention-based text recognition models, the decoder recurrently outputs predictions.
Specifically, the prediction of the previous step is often embedded into high-dimensional feature space, and the embedded vector will directly participate in the next decoding step as a guidance.
Note that the intensity of blue indicates the weighting value of guidance.
As shown in Figure~\ref{Figure_Compare} (b), changeless intensity of blue represents that all the guidance weights are invariant in traditional attention-based models, regardless of correlation between the previous and the current prediction.

\begin{figure}[t]
\centering
\setlength{\leftskip}{-5pt}
\includegraphics[width=0.5\textwidth]{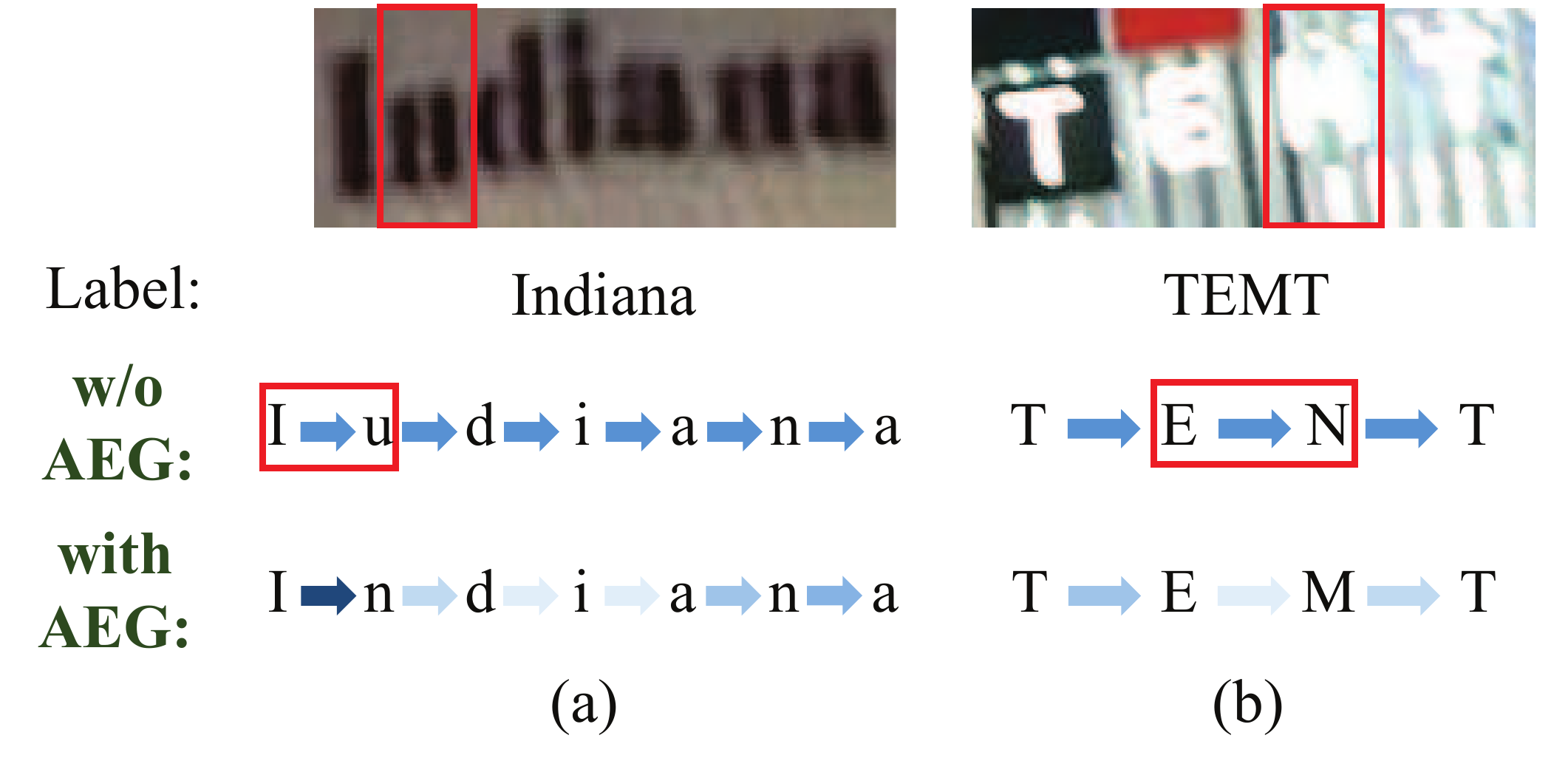}
\caption{Illustration for the guidance of previous prediction in attentional decoding stage without/with the proposed AEG.
An arrow with deeper color corresponds to a larger weighting value of the guidance.
(a): A normal word.
(b): A meaningless string.
}
\label{Figure_motivation}
\end{figure}

According to character language modeling \cite{bengio2003neural}, \cite{islam2012comparing}, \cite{kim2016character},  character correlations can be reflected in high order statistics, e.g., the higher co-occurrences probability indicates stronger correlation of the neighboring characters.
As shown in Figure~\ref{Figure_motivation} (a), the pair of ``In" in the word ``Indiana" is a ``strong-correlated" pair because that it appears frequently in common-used words, while the pair of ``ia" is a ``weak-correlated" pair.
Similarly, the pairs of ``te" and ``em" in Figure~\ref{Figure_motivation} (b) are strong-correlated and weak-correlated respectively.

As illustrated in Figure~\ref{Figure_motivation} (a), when it comes a word ``Indiana", it is more suitable to decode the second character ``n" rather than ``u" with the guidance of previous prediction ``I".
However, as shown in Figure~\ref{Figure_motivation} (b), when it comes a meaningless string `temt', the wrong ``strong-correlated" guidance (marked with red box) of previous prediction ``e"  misleads the existing attentional text recognizer and results in decoding errors ``en".
Therefore, the invariant weight of guidance is inappropriate and may be harmful. 

Motivated by the observations above, it is reasonable to introduce high-order character language model to attention mechanism for a proper guidance. 

\subsection{Adaptive Embedding Gate (AEG)}
In this paper, we propose a new module for attention-based scene text recognizer, namely \textbf{adaptive embedding gate} (\textbf{AEG} in short).
As illustrated in Figure~\ref{Figure_Compare}(c), AEG focuses on adaptively estimating the correlations between adjacent characters by controlling transmission weight of the previous embedded vector.
Specifically, AEG can strengthen the correlation of strong-correlated pair while weaken the guidance weight within weak-correlated pair.
As shown in Figure~\ref{Figure_motivation}, AEG selectively apply different weight of guidance for different character-pairs and correct the recognition results eventually.
Further, the proposed AEG is a flexible module that can be easily integrated with existing attentional methods \cite{shi2018aster}, \cite{cluo2019moran} to improve the performance in an end-to-end manner.

Our primary contributions are summarized as follows:
\begin{itemize} 
\item 
We explore the existing attention mechanism for scene text recognition and point out that the inappropriate use of previous prediction restricts the recognition performance in decoding stage.
\item 
We propose a novel module called AEG to introduce proper correlations between characters.
Further, the formulation and three implementations of AEG are introduced in this paper.
\item 
Extensive experiments are conducted on various scene text benchmarks, demonstrating the performance superiority and flexibility of AEG.
\item 
The proposed AEG significantly improve the robustness of the existing attention mechanism under different noise disturbances, e.g., Gaussian blur, salt and pepper noise and random occlusion.
\end{itemize}

The reminder of this paper is organized as follows.
Section~\ref{related_work} gives a brief review of related work on scene text recognition.
Section~\ref{methodology} introduces the details of our proposed module AEG.
Section~\ref{experiments} evaluates the proposed approach on various benchmark datasets.
Finally, Section~\ref{conclusion} concludes the paper.

\begin{figure*}[th]
\centering
\includegraphics[width=0.85\textwidth]{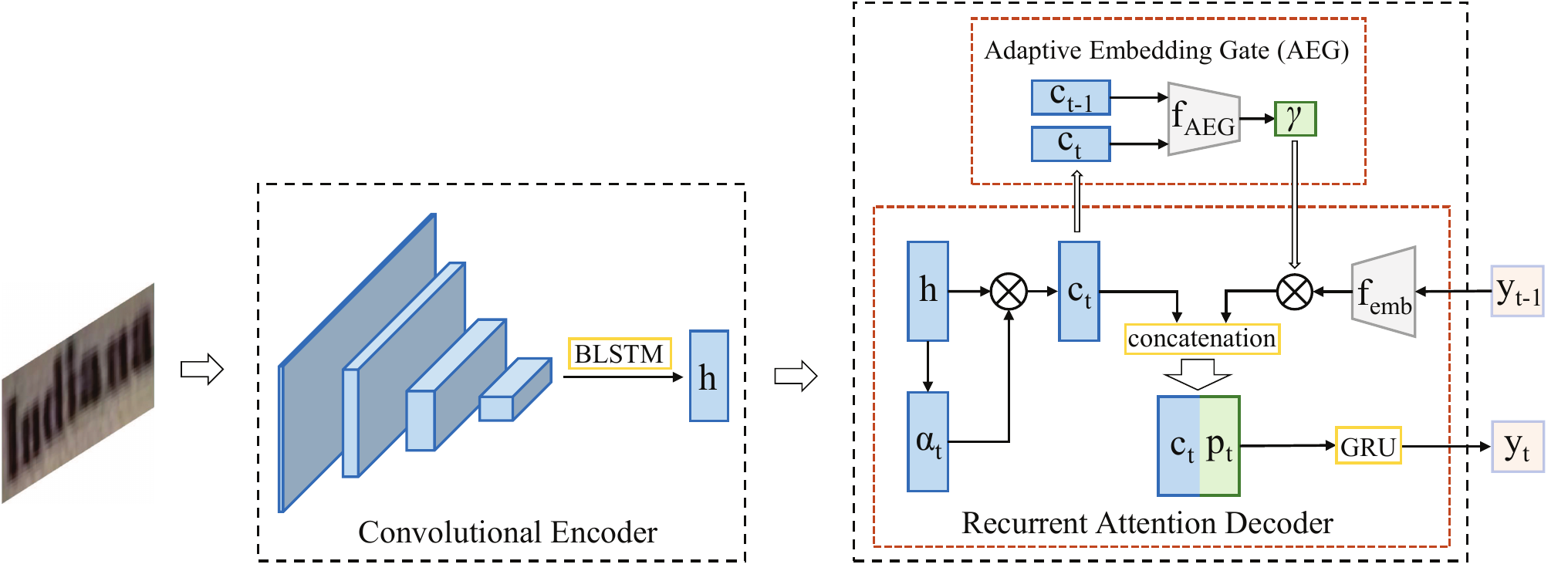}
\caption{Overall architecture of attention-based text recognizer with the proposed AEG.}
\label{Figure_overview}
\end{figure*}

\section{Related Work}
\label{related_work}

Scene text recognition has attracted significant research interest in the computer vision community\cite{jo2015led}, \cite{lee2016recursive}, \cite{sain2018multi}, \cite{shi2018aster}. 
Comprehensive surveys can be found in \cite{ye2014text}, \cite{zhu2016scene}, \cite{long2018scene}.

Early work relies on low-level features for scene text recognition, such as histogram of oriented gradients descriptors \cite{wang2012end}, connected components \cite{neumann2012real}, stroke width transform \cite{yao2014unified} etc.
However, the performances of these methods are limited by the low capability of features. 
With the rapid development in neural networks, a large number of effective frameworks have emerged in scene text recognition. 
These methods can be divided into two branches. 

One branch is based on segmentation.
It attempts to locate the position of each character from the input text image, and applies a character classifier to obtain the recognition results.
For instance, Bissacco et al. \cite{bissacco2013photoocr} proposed a neural network with five hidden layers for character recognition and used an n-gram approach for language modeling. 
Wang et al. \cite{wang2012end} used a CNN to recognize characters and adopt a non-maximum suppression to obtain the final predictions. 
Jaderberg et al. \cite{jaderberg2014deep} proposed a weight-shared CNN for unconstrained text recognition.
All the aforementioned pipelines require the accurate individual detection of characters, therefore the quality of the character detectors limits the recognition performance.

The other branch is segmentation-free.
It recognizes the text line as a whole and focuses on mapping the entire image to a word string directly by exploiting CTC-based algorithm or attention mechanism.
For instance, the CTC loss was often combined with the RNN outputs for calculating the conditional probability between the predicted and the target sequences in \cite{liu2016star}, \cite{shi2017end}, \cite{liu2018squeezedtext}.
Recently, an increasing number of recognition approaches based on the attention mechanism have achieved significant improvements \cite{cluo2019moran}, \cite{shi2018aster}, \cite{shi2016robust}, \cite{he2019vd}, \cite{cheng2017focusing}.
However, as discussed in Section~\ref{introduction}, in the existing attentional decoding mechanism the weight of previous prediction guidance are invariant, which is inappropriate and may be harmful.
Therefore, we propose AEG to introduce the proper guidance for attentional scene text recognizer.

The method we propose derives from the idea of the character language modeling\cite{bengio2003neural}, \cite{islam2012comparing}, \cite{kim2016character}, where character correlations can be reflected in high order statistics.
For instance, Marti et al. \cite{marti2001using} used a statistical language model to improve the performance of handwriting recognition.
Li et al. \cite{li2006vector} applied the statistics of letters for spoken language identification.
Islam et al. \cite{islam2012comparing} measured word relatedness based on co-occurrence statistics.
Specifically, we focus on introducing high-order character language models to existing attentional decoding stage for a proper guidance. 

As our proposed AEG is a refined version of the existing attention mechanism, it reasonably falls into the segmentation-free category.
The overview of our model is shown in Figure~\ref{Figure_overview}, and we will detail the method in Section~\ref{methodology}.

\section{Methodology}
\label{methodology}

As an overview of our proposed model shown in Figure~\ref{Figure_overview}, our model consists of two components: 
1) a convolutional encoder network that extracts features from an input image and converts features to high-level visual representations. 
2) a recurrent attention-based decoder network that combined with the proposed AEG to generate target sequences.
In the following sections, we first describe the two components in Section~\ref{convolutional_encoder_network} and Section~\ref{recurrent_attention_decoder_network} respectively.
Then, we introduce the formulation description and instance implementations of the proposed AEG in Section~\ref{adaptive_embedding_gate}.

\subsection{Convolutional Encoder Network}
\label{convolutional_encoder_network}

Scene text recognition aims at directly translating a visual image $I$ into a target string sequence.
Therefore, rich and discriminative features are critical to recognition performance.

A ResNet-based \cite{he2016deep} feature extractor is adopted as the primary structure for the convolutional encoder network.
The encoder first extracts a feature map from an input image $I$.
However, features extracted by CNN are constrained by their receptive fields.
To enlarge the image region for feature expressions, we employ a two-layer Bidirectional Long Short Term Memory (BLSTM) network \cite{graves2008novel} over the feature map.
The encoding process is represented as follows:
\begin{equation}
F_{e}(I)=(h_{1},h_{2}...h_{N}),
\label{feature_equation}
\end{equation}
where $N$ is the length of extracted feature sequence.

\subsection{Recurrent Attention Decoder Network}
\label{recurrent_attention_decoder_network}

The recurrent decoder network aims at translating the encoded features into the prediction sequence, where the attention mechanism is used to align the prediction sequence $y = (y_{1},y_{2}...y_{T})$ and the ground truth $g = (g_{1},g_{2}...g_{T})$.
$T$ indicates the maximum decoding step size.

At the $t$-th step, the recognition model generates an output $y_{t}$,
\begin{equation}
y_t=Softmax(W_{o}s_t+b_{o}),
\end{equation}
where $s_{t}$ is the hidden state of Gated Recurrent Unit (GRU) \cite{cho2014properties} at time $t$.
Specifically, the GRU is a variation of the RNN, typically used to model long-term dependencies. 
Further, $s_{t}$ is computed as
\begin{equation}
\label{cat_equation}
s_t = GRU([f_{emb}(y_{t-1}), c_t], s_{t-1}),
\end{equation}
where $[f_{emb}(y_{t-1}), c_t]$ is the concatenation of $f_{emb}(y_{t-1})$ and $c_{t}$. 
$f_{emb}(y_{t-1})$ denotes the embedding vectors of the previous prediction $y_{t-1}$.
We adapt a one-dimensional attention mechanism, where $c_{t}$ is the relevant contents computed as the weighted sum of features,
\begin{equation}
c_t =\sum^N_{j=1}\alpha_{t, j}h_j.
\end{equation}
$N$ represents the feature length, which is same as that in Equation.~\ref{feature_equation}. $\alpha_{t, j}$ is the vector of attention weights, $\alpha_{t} \in R^{N}$, expressed as follows:
\begin{equation}
\alpha_{t, j}=\frac{exp(e_{t, j})}{\sum^N_{i=1}exp(e_{t, i})},
\end{equation}
where $e_{t, j}$ is the alignment score which represents the degree of correlation between the high-level feature representation and the current output,
\begin{equation}
e_{t, j}=f_{attn}(s_{t-1}, h_j).
\end{equation}
The alignment function $f_{attn}$ is parameterized by a single-layer multilayer perceptron, such that
\begin{equation}
f_{attn}(s_{t-1}, h_j)=V_aTanh(W_{s}s_{t-1}+W_{f}h_{j} + b_{a}).
\end{equation}
In the above, $W_{o}$, $b_{o}$, $V_{a}$, $W_{s}$, $W_{f}$, and $b_{a}$ are trainable parameters. 

The decoder completes the generation of characters when it predicts an end-of-sequence token ``$EOS$." \cite{sutskever2014sequence} 
We optimize the parameters by minimize the loss function of the recurrent attention decoder network as follows:
\begin{equation}
\label{attn_equation}
L_{attn} = -\sum^T_{t=1}log P(g_t|I,\theta),
\end{equation}
where $\theta$ is the parameters of the network.

\subsection{Adaptive Embedding Gate}
\label{adaptive_embedding_gate}

For a proper guidance of previous prediction, we propose AEG to adaptively estimate the correlations between adjacent characters.
We first introduce the general formulation description of AEG and then we provide several specific instance implementations of it.

\subsubsection{Formulation}
\label{formulation}

AEG apply a proper guidance by controlling transmission weight of the previous embedded vector.
Therefore, we redefine the Equation.~\ref{cat_equation} as:
\begin{equation}
\label{cat_equation_redefine}
s_t = GRU([p_t, c_t], s_{t-1}),
\end{equation}
where $p_t$ is the output vector of AEG.
Given the relevant contents $c_t$ and the previous prediction $y_{t-1}$, $p_t$ is expressed as follows:
\begin{equation}
p_{t} = f_{AEG}(c_t,c_{t-1}) f_{emb}(y_{t-1})
\end{equation}
The pairwise function $f_{AEG}$ computes a scalar called AEG score, which reflects the degree of correlations and decides the weight of guidance in the next step.
The AEG score ranges from $0$ to $1$, the greater value implies the stronger correlation and vice versa.
The unary function $f_{emb}$ is the same with that of Equation.~\ref{cat_equation}, aiming at embedding the previous prediction $y_{t-1}$ into high-dimensional feature space.

AEG is a flexible and robust building block and can be easily incorporated together with the existing attention-based method.
Furthermore, AEG has diverse implementations to adaptively estimate the correlations.

\subsubsection{Instantiations}
\label{instantiations}

Next we describe several versions of $f_{AEG}$.
The output of $f_{AEG}$, i.e., the AEG score, can be adaptively computed by the neighboring contents, and the corresponding experimental results will be shown in Table~\ref{Table_instantiations} , Section~\ref{experiments_on_instantiations}

\begin{itemize} 

\item
\textbf{Add.}

Following the alignment model \cite{bahdanau2014neural} which scores the degree of two variables matching, a natural implementation of $f_{AEG}$ is to directly add the corresponding elements of neighboring contents.
In this paper we consider:
\begin{equation}
f_{AEG_{Add}}(c_{t},c_{t-1})=\sigma(V_{c}Tanh(W_{p}c_{t-1}+W_{c}c_t + b_{c})).
\end{equation}
Here $\sigma$ represents the sigmoid function and $Tanh$ is the activation function.
Further, $V_{c}$, $W_{p}$, $W_{c}$, and $b_{c}$ are all trainable parameters.

\item
\textbf{Dot product.}

Dot product is implementation-friendly in modern deep learning platforms.
Besides add version, we also evaluate a dot-product form of $f_{AEG}$.
Specifically, $f_{AEG}$ can also be defined as a dot-product similarity:
\begin{equation}
f_{AEG_{Dot}}(c_{t},c_{t-1})=\sigma((W'_{c}c_{t})^T(W'_{p}c_{t-1})).
\end{equation}
Here $W'_{c}$, $W'_{p}$ are all weight matrix to be learned.

\item
\textbf{Concatenation.}

Concatenation is used by the pairwise function in Relation Networks \cite{santoro2017simple} for visual reasoning.
In this paper we consider a concatenation form of $f_{AEG}$:
\begin{equation}
f_{AEG_{Concat}}(c_{t},c_{t-1})=\sigma(V''_{c}Tanh([W''_{c}c_t,W''_{p}c_{t-1}]+b''_{c})),
\end{equation}
where $[\cdot,\cdot]$ denotes concatenation of $c_t$ and $c_{t-1}$.
Similarly, $V''_{c}$, $W''_{p}$, $W''_{c}$, and $b''_{c}$ are all trainable parameters.
\end{itemize}

The above several variants demonstrate the flexibility of our proposed AEG.
However, the implementation of AEG is not limited to these.
We believe alternative versions are possible and may further improve recognition performance for scene text.

\subsection{AEG Training}
\label{aeg_training}

To guide AEG constructing appropriate correlations between adjacent characters, we design several versions of training mechanism.
We will show experimental results by Table~\ref{Table_supervision}, Section~\ref{experiments}, and analyze the effect of these variants in details.

For simplicity, the vector consisting of AEG scores is recorded as $\gamma$, $\gamma \in R^{T}$.
Given an input image $I$ and its ground truth string $g = (g_{1},g_{2}...g_{T})$, we construct the labeling for vector $\gamma$ as $\gamma^{gt}$ of the same length $T$.

\begin{algorithm}[h] 
\caption{Word-Frequency-based $\gamma^{gt}$ labeling} 
\begin{algorithmic}[1] 
\label{algorithm_dictionary}
\REQUIRE ~~\\
  \hspace*{0.02in} {\bf Input:} 
  ground truth $g$, transition probability matrix $M$;\\
  \hspace*{0.02in} {\bf Output:} 
  labeling vector of AEG score $\gamma^{gt};$
 
\STATE Initialize $\gamma^{gt} = zeros((T))$
\FOR{$l \in [1, 2, ..., T]$} 
\IF{$l = 1$}
\STATE $\gamma^{gt}[t] = 0$
\ELSE
\STATE $\gamma^{gt}[t] = M[g[t - 1]][g[t]]$
\ENDIF
\ENDFOR
\STATE Return $\gamma^{gt}$
\end{algorithmic} 
\end{algorithm}

\subsubsection{Training with Word Frequency}
\label{training_with_word_frequency}

To capture universal character-pair correlations, a natural choice is applying statistics about word frequency\cite{islam2012comparing}, i.e., ``strong-correlated" pairs means that the pairs frequently appear in words and vise versa.
Therefore, we adapt a dictionary\footnote{\url{https://www.oxfordwordlist.com/pages/report.asp}} with $9121$ commonly used words.
The duplicates and single characters have been removed.
Further, we construct $\gamma^{gt}$ as follows:

First, we count the frequencies of adjacent character-pairs in the dictionary, which are normalized to get a $26\times26$ transition probability matrix $M$.
The values in $M$ denotes the transition probabilities between $26$ letters.
Specifically, greater value corresponds to the stronger correlations between character-pairs.
Then, as illustrated in Algorithm~\ref{algorithm_dictionary}, $\gamma^{gt}_{t}$ equals to the transition probabilities of specific character-pairs in matrix, i.e., the probabilities between the ground truth $g_{t-1}$ and $g_t$, $1 \leq t \leq T$.
In particular, the transition probabilities between two digits or between a digit and a character is set to $0$.

\begin{algorithm}[h] 
\caption{Root-based $\gamma^{gt}$ labeling} 
\begin{algorithmic}[1] 
\label{algorithm_root} 
\REQUIRE ~~\\
  \hspace*{0.02in} {\bf Input:} 
  ground truth $g$, root table $R$;\\
  \hspace*{0.02in} {\bf Output:} 
  labeling vector of AEG score $\gamma^{gt};$
\STATE Initialize $\gamma^{gt} = zeros((T)))$
\FOR{each root $r_i \in R$} 
\IF{root $r_i \in g$}
\STATE denotes the index of the first/last character of $r_i$ in $g$ as $S$/$E$
\STATE $\gamma^{gt}[S:E] = \gamma^{gt}[S:E] + 1$
\ENDIF
\ENDFOR
\STATE $\gamma^{gt} = normalized (\gamma^{gt}) $
\STATE Return $\gamma^{gt}$
\end{algorithmic} 
\end{algorithm}

\subsubsection{Training with root}
\label{training_with_root}

Besides the word frequency, we also consider applying statistics about root.
The character-pair which constitutes a root means ``strong-correlated" and vise versa. 
Therefore, we adapt a root table\footnote{\url{https://www.quia.com/files/quia/users/skrichard/ComSkills2/RootsPrefixesSuffixes}}. 
The root table includes $707$ typically used roots, and has been removed duplicates and single character.
Specifically, the length of the roots is distributed between $2$ and $10$ characters. 
Among them, the roots of $3-4$ character-length, e.g., ``ing" and ``ance", constitute the largest proportion, i.e., approximately $71.99\%$, while few roots are longer than eight characters.
We construct $\gamma^{gt}$ as follows: 

As demonstrated in Algorithm~\ref{algorithm_root}, the labeling $\gamma^{gt}$ is initialized to a zero vector.
If two adjacent characters comprise a root, the corresponding position of $\gamma^{gt}$ is increased by one. 
Similarly, the value corresponding to non-root or digits is set to $0$.
During training, $\gamma^{gt}$ is normalized, ensuring that the range of the value is between $0$ and $1$.

\subsubsection{Training with weakly supervision}
\label{training_with_weakly_supervision}

We consider that whether the network can autonomously learn the correlations between adjacent characters with weakly supervised learning strategy.
No direct supervision information is provided to $\gamma$.
The learning of $\gamma$ relies entirely on scene text recognition task.

\subsection{Training Loss}
\label{training_loss}

We define a AEG loss $L_{AEG}$ conditioned on $\gamma$ and $\gamma^{gt}$, to penalize the attention when it does not obtain the correct correlations between adjacent characters. 
$L_{AEG}$ is computed as follows:
\begin{equation}
L_{AEG} = MSELoss(\gamma , \gamma^{gt}) = \frac{1}{T}\sum^{T}_{t=1}(\gamma_{t}-\gamma_{t}^{gt})^2,
\end{equation}
where $MSELoss$ indicates the mean square value of the difference between the predicted value and ground truth.

The final optimization objective $L$ is the weighted sum of $L_{AEG}$ and $L_{attn}$.
Specifically, $L_{attn}$ is introduced in Equation.~\ref{attn_equation}.
Therefore, $L$ is formulated as,
\begin{equation}
L = L_{attn} + \lambda L_{AEG}.
\end{equation}
The hyper-parameter $\lambda$ is introduced to balance the two terms.
We set $\lambda$ to $1$ in our experiments.

\section{Experiments}
\label{experiments}

In this section, we systematically verify the effectiveness and robustness of the proposed AEG. 
Extensive experiments were conducted on a number of benchmarks for scene text recognition, including the IIIT$5$K, SVT, SVT-P, CUTE$80$, and ICDAR datasets.
Experimental results demonstrate the performance superiority of our method. 

Specifically, we begin by specifying the experimental settings in Section~\ref{experimental_settings}.
Then we conduct a few ablation studies in Section~\ref{experiments_on_instantiations}, Section~\ref{experiments_with_different_supervision} and Section~\ref{effect_of_previous_embedding}, each aims at demonstrating its effectiveness and analyzing its behavior.
Finally, in Section~\ref{comparison_to_state_of_the_art} and Section~\ref{robust_of_AEG}, we compare AEG-based methods with other state-of-the-arts and show the outstanding robustness of AEG under different noise disturbances.

\subsection{Experimental Settings}
\label{experimental_settings}

\subsubsection{Datasets}
\label{datasets}

\textbf{IIIT5k-Words (IIIT5K)} \cite{mishra2012scene}: is collected from the Internet and contains $3000$ cropped word images for testing.
It provides a $50$-word and a $1000$-word lexicons for each image in the dataset.

\textbf{Street View Text (SVT)} \cite{wang2011end}: is collected from the Google Street View, containing $647$ word images for testing.
It provides a $50$-word lexicon for each image in the dataset. Many images are corrupted by noise and blur.

\textbf{ICDAR 2003 (IC03)} \cite{lucas2005icdar}: contains $251$ scene text images. It provides a $50$-word lexicon defined by Wang et al. \cite{wang2011end} and a ``full-lexicon" for each image. 
For a fair comparison, we discard images that contain non-alphanumeric characters or those with less than three characters, following Wang, Babenko, and Belongie \cite{wang2011end}.
The resulting dataset consists $867$ cropped images.

\textbf{ICDAR 2013 (IC13)} \cite{karatzas2013icdar}: inherits most images from IC$03$ and extends it with some new images. 
It consists of $1015$ cropped images without an associated lexicon.

\textbf{SVT-Perspective (SVT-P)} \cite{neumann2012real}: is collected from the side-view angle snapshots in Google Street View that contains $639$ cropped images for testing. 
It provides a $50$-word lexicon and a ``full-lexicon."
Most of the images are heavily distorted.

\textbf{CUTE80 (CUTE)} \cite{risnumawan2014robust}: focuses on curved texts, consisting of $80$ high-resolution images captured in natural scenes.
This dataset contains $288$ cropped natural images for testing without an associated lexicon.

\textbf{ICDAR 2015 Incidental Text (IC15)} \cite{karatzas2015icdar}: contains $2077$ cropped images.
A large proportion of images are blurred and multi-oriented. No lexicon is associated.

\textbf{Synth90k} \cite{jaderberg2014synthetic}: contains $8$ million synthetic images of cropped word generated from a set of $90$k common English words.
Words are rendered onto natural images with random transformations and effects.
Every image in Synth$90$k is annotated with a ground truth word.

\textbf{SynthText} \cite{gupta2016synthetic}: contains $6$ million synthetic images of cropped word.
The generation process is similar to that of Synth$90$k.

\subsubsection{Network}
\label{Network}

\begin{figure}[t]
\centering
\includegraphics[width=0.4\textwidth]{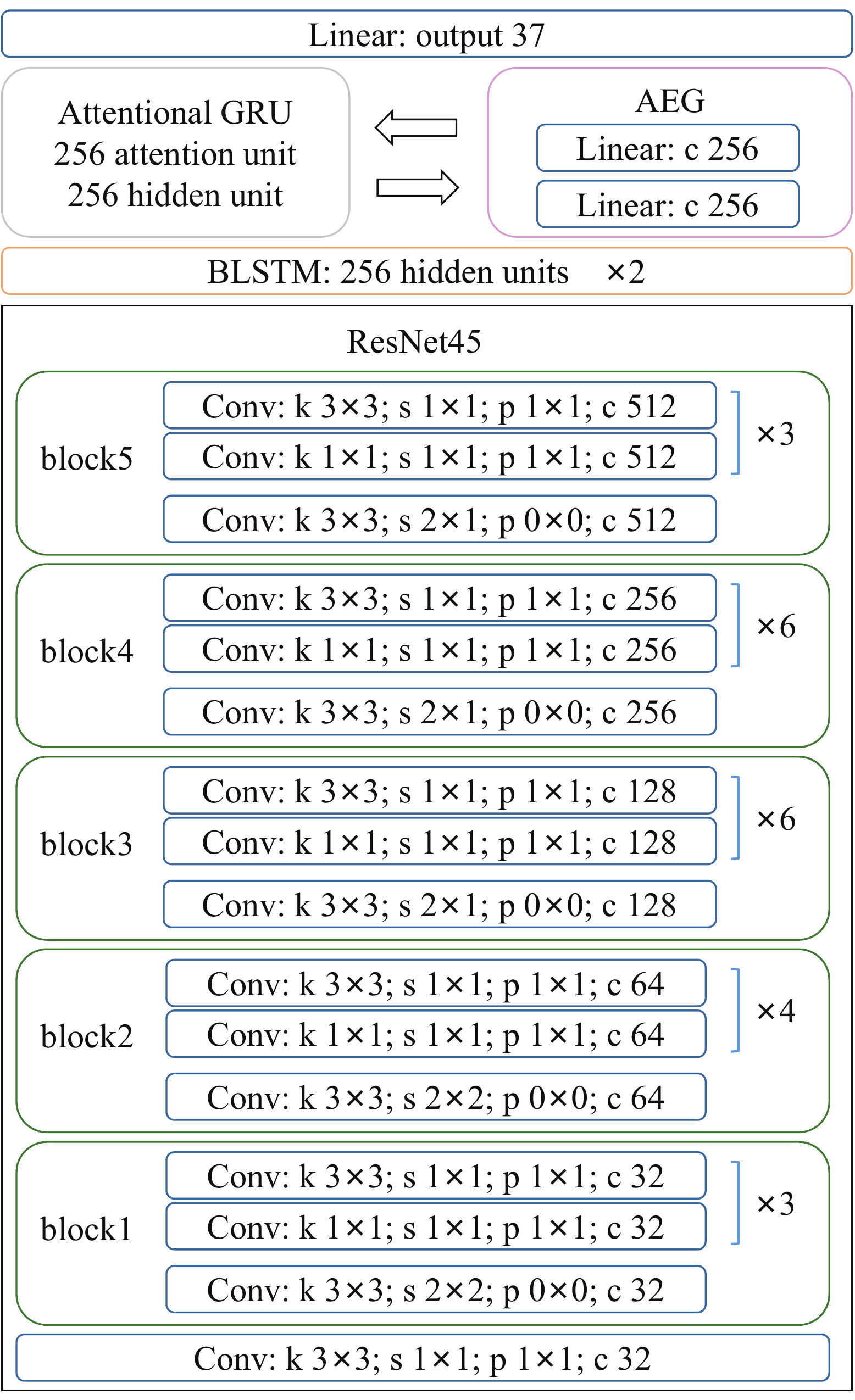}
\caption{
Illustration of text recognition network architecture.
``k", ``s", ``p" and ``c" stands for the kernel, stride, padding and channel size of the convolutional layer in a block, respectively.
``Attentional GRU" means attention-based GRU decoder and ``AEG" represents the proposed adaptive embedding gate.
}
\label{Figure_network}
\end{figure}

All the input images are resized to $32 \times 100$.
Details about the experimental model is given in Figure~\ref{Figure_network}.
For the encoder, we use a residual network with five blocks to extract features from the input image.
Following the residual network are two-layers of BSLTM with $256$ hidden states.
As illustrated in Figure~\ref{Figure_network}, the decoder is based on the attention mechanism combined with the proposed AEG.
The number of attention units and hidden units of the GRU are both $256$.
Specifically, ``Baseline" in the following sections indicates the attentional text recognizer without AEG.
The number of output categories is $37$, including $26$ letters, $10$ digits, and a symbol representing ``$EOS$."

\subsubsection{Optimization}
\label{optimization}

With the ADADELTA \cite{zeiler2012adadelta} optimization method, we train our model entirely on synthetic images of Synth$90$k and the SynthText from scratch.
No extra data is used.
Further, for fair comparison, no pre-trained model is used during training, and no further fine-tuning on any test datasets.
The learning rate is set as $1.0$ initially and decay to 0.01 at step $0.8$M.
Following the settings in \cite{shi2018aster}, we find the classic learning rate schedule beneficial to performance.
The batch size is set to $64$ in both the training and testing.

\subsubsection{Implementation}
\label{implementation}

All of our experiments are performed under the Pytorch \cite{Pytorch} framework.
CUDA $8.0$ and CuDNN v$7$ backend are used in our experiments.
The proposed model is trained on a single NVIDIA GTX-$1080$Ti graphics card with $11$GB memory.
The training speed is about 5 iterations/s, taking less than 2 days to reach convergence.
Our method takes about $9.8$ms to recognize an image.

\subsection{Experiments on Instantiations}
\label{experiments_on_instantiations}

\begin{table}[htbp]
\scriptsize
  \centering
  \caption{Performance comparison among three instantiations of AEG in attention-based recognition model.}
    \setlength{\leftskip}{-3pt}
    \begin{tabular}{ccccccccc}
    \toprule
    \textbf{Exp} & \textbf{Variants} & \textbf{IIIT5k} & \textbf{SVT} & \textbf{IC03} & \textbf{IC13} & \textbf{SVT-P} & \textbf{CUTE} & \textbf{IC15} \\ 
    \midrule
    (a) & Baseline & 92.6  & 87.5  & 93.6  & 92.2  & 76.2  & 78.2  & 71.6 \\ 
    (b) & Add  & \textbf{93.6} & \textbf{89.2} & \textbf{94.8} & 92.9  & 80.0    & \textbf{80.2} & \textbf{75.5} \\ 
    (c) & Dot product & 93.2  & 88.4  & 94.1  & \textbf{93.7} & \textbf{80.8} & 78.5  & 73.6 \\ 
    (d) & Concatenation & 93.2  & 88.5  & 94.2  & 91.6  & 80.0    & 77.1  & 74.4 \\ 
    \bottomrule
    \end{tabular}%
  \label{Table_instantiations}%
\end{table}%

Table~\ref{Table_instantiations} compares different instantiations of $f_{AEG}$ added to the AEG-based attentional encoder-decoder recognition model.

Most of instantiations can lead to obvious improvement over the baseline except ``Concatenation" on IC$13$ and CUTE, indicating that the generic AEG behavior is the main reason for the observed improvements.

Furthermore, as illustrated in experiments (b), Table~\ref{Table_instantiations}, the ``Add" version of $f_{AEG}$ achieves the best recognition performance, demonstrating that correlations between adjacent characters is easier to be captured in this way.

In the rest of this paper, we use the add version of $f_{AEG}$ by default.

\subsection{Experiments with different supervision}
\label{experiments_with_different_supervision}

\begin{table}[htbp]
\scriptsize
  \centering
  \caption{Performance comparison among three supervisions of AEG in attention-based recognition model.}
  \setlength{\leftskip}{-3pt}
    \begin{tabular}{ccccccccc}
    \toprule
    \textbf{Exp} & \textbf{Variants} & \textbf{IIIT5k} & \textbf{SVT} & \textbf{IC03} & \textbf{IC13} & \textbf{SVT-P} & \textbf{CUTE} & \textbf{IC15} \\
    \midrule
    (a) & Baseline & 92.6  & 87.5  & 93.6  & 92.2  & 76.2  & 78.2  & 71.6 \\ 
    (b) & Weakly & 92.8    & 88.6  & 93.9  & 92.5  & 77.0  & 78.5  & 72.1 \\
    (c) & Word Frequency & 93.4  & \textbf{90.0} & 94.3  & 92.7  & \textbf{80.4} & 79.9  & 74.7 \\
    (d) & Root table & \textbf{93.6} & 89.2  & \textbf{94.8} & \textbf{92.9} & 80.0    & \textbf{80.2} & \textbf{75.5} \\   
    \bottomrule
    \end{tabular}%
  \label{Table_supervision}%
\end{table}%

Table~\ref{Table_supervision} compares the experimental results with different supervision.

It is clear that all these variants can learn some certain correlations between adjacent characters.
Furthermore, direct supervision strategy is better than weakly supervision of recognition task, i.e., the performance of (c) or (d) is better than experiment (b).

Interestingly, the word frequency and root table versions perform similarly, up to some random variations on the test benchmark datasets.
It shows that our AEG are not sensitive to these direct supervision choices, indicating that the supervised behavior of AEG is not the key to the improvement in our applications; instead it is more likely that the proper guidance of previous prediction is important.

In the following experiments, we just use the root table as the guidance of previous prediction by default.

\subsection{Experiments with different previous prediction}
\label{effect_of_previous_embedding}

\begin{table}[htbp]
\scriptsize
\centering
\caption{Performance comparison among four variations of previous prediction (short in ``pre") in attention-based recognition model.}
\setlength{\leftskip}{-18pt}
\begin{tabular}{ccccccccc}
\toprule
\textbf{Exp}    &  \textbf{Method}           & \textbf{IIIT5k} & \textbf{SVT}   & \textbf{IC03}  & \textbf{IC13}  & \textbf{SVT-P} & \textbf{CUTE} & \textbf{IC15}  \\ \midrule
(a) & Baseline & 92.6  & 87.5  & 93.6  & 92.2  & 76.2  & 78.2  & 71.6 \\ 
(b)&No pre   & 89.9  & 85.4 & 92.7 & 91.1 & 74.1 & 75.0 & 68.5 \\ 
(c)&Random pre      & 89.4  & 83.6 & 91.3 & 89.4 & 70.4 & 73.6 & 66.6 \\ 
(d)&Baseline + AEG   & \textbf{93.6}  & \textbf{89.2} & \textbf{94.8} & \textbf{92.9} & \textbf{80.0} & \textbf{80.2}  & \textbf{75.5} \\ 
\bottomrule
\end{tabular}
\label{Table_prior}
\end{table}

Table~\ref{Table_prior} compares the experimental results on different strategies of using previous prediction.
``No pre" indicates the experiments which are based on the attentional model without previous prediction, while ``Random pre" represents that the previous prediction is randomly selected from $37$ output categories.

To explore the role of previous prediction, we remove the $p_{t}$ in Equation.~\ref{cat_equation_redefine}.
The corresponding experimental results are recorded as ``No pre" in Table~\ref{Table_prior}.
By comparing experiments (a) and (b), the attentional model without previous prediction exhibits an apparent decline, which verifies the validity of the previous prediction.

Moreover, the performance of the experiment (c) is significantly worse than others.
Obviously, random selected previous prediction provides completely confusing correlations between adjacent characters and eventually influences text recognition, which indicates that the increase in parameters is not the cause of performance improvement.
Therefore, inappropriate guidance of previous prediction may be harmful to the network.

In experiment (d), the proposed AEG-based attentional model outperforms baseline on all test datasets.
As mentioned in Section~\ref{introduction}, the inappropriate use of previous predictions in existing attention mechanism restricts the recognition performance, while AEG provides more appropriate guidance by adaptively estimating the correlations between adjacent characters.

By comparing experiments (c) and (d), the importance of applying appropriate previous prediction guidance in a recognition model is self-evident.

\subsection{Comparison to State-of-the-Arts}
\label{comparison_to_state_of_the_art}

Finally, we compare the performance of our proposed AEG-based model with other state-of-the-art models.
When a lexicon is given, we simply replace the predicted word with nearest lexicon word under the metric of edit distance.

\subsubsection{Results on Regular Benchmarks}
\label{results_on_regular_benchmarks}

In regular benchmarks, most of testing samples are horizontal text and a small part of them are distorted text.

As illustrated in Table~\ref{Table_regular}, AEG significantly boost the recognition performance of baseline, which indicates the proper guidance of previous prediction is important.
Besides, baseline with AEG achieves comparable results to the state-of-the-art models.

Further, AEG is a flexible building block and can be easily used together with the existing attentional method.
With structures of ASTER\cite{shi2018aster} and MORAN-v$2$\cite{cluo2019moran}, the AEG-based model significantly outperform all current state-of-the-art methods in lexicon-free mode.
In particular, on IC$13$ and SVT, incorporating with AEG, the performance of ASTER and MORAN-v$2$ have been improved by $1.4\%$ and $3.2\%$ respectively.

\subsubsection{Results on Irregular Benchmarks}
\label{results_on_irregular_benchmarks}

\begin{table}[b]
\caption{Performance comparison on irregular benchmarks. ``$50$" and ``$1$k" are lexicon sizes. ``Full" indicates the combined lexicon of all images in the benchmarks. ``None" means lexicon-free. ``*" represents scene text recognition methods with rectification. ``$^\dagger$" indicates the scene text recognition methods with extra datasets.}
\label{Table_irregular}
\centering
\scriptsize
    \begin{tabular}{c|ccc|c|c}
    \toprule
    \multirow{2}[4]{*}{Method} & \multicolumn{3}{c|}{SVT-P} & CUTE  & IC15 \\
\cmidrule{2-6}          & 50  & Full  & None  & None  & None \\
    \midrule
    ABBYY \cite{wang2011end} & 40.5  & 26.1  & -     & -     & - \\
    Mishra et al. \cite{mishra2012top} & 45.7  & 24.7  & -     & -     & - \\
    Wang et al. \cite{wang2012end} & 40.2  & 32.4  & -     & -     & - \\
    *Shi et al. \cite{shi2016robust} & 91.2  & 77.4  & 71.8  & 59.2  & - \\
    *Liu et al. \cite{liu2016star} & 94.3  & 83.6  & 73.5  & -     & - \\
    $^\dagger$Yang et al. \cite{yang2017learning} & 93.0  & 80.2  & 75.8  & 69.3  & - \\
    Cheng et al. \cite{cheng2018aon}  & 94.0  & 83.7  & 73.0    & 76.8  & 68.2 \\
    *Liu et al. \cite{liu2018char} & -     & -     & 73.5  & -     & 60.0 \\
    *Zhan et al. \cite{zhan2019esir} & -     & -     & 79.6  & \textbf{83.3}  & 76.9 \\
    *Luo et al. \cite{cluo2019moran} & 94.3  & 86.7  & 76.1  & 77.4  & 68.8 \\
    \midrule
    Baseline & 91.3  & 84.6  & 76.2  & 78.2  & 71.6 \\
    Baseline + AEG(ours) & 92.6  & 86.4  & $80.0^{\color{red}\uparrow3.8}$  & $80.2^{\color{red}\uparrow2.0}$  & $75.5^{\color{red}\uparrow3.9}$ \\
    \midrule
    *ASTER \cite{shi2018aster} & -     & -     & 78.5  & 79.5  & 76.1 \\
    ASTER + AEG(ours) & 94.4  & 89.5  & $82.0^{\color{red}\uparrow3.5}$  & $80.9^{\color{red}\uparrow1.4}$  & $76.7^{\color{red}\uparrow0.6}$ \\
    \midrule
    *MORAN-v2 \cite{cluo2019moran} & 94.4     & 88.3     & 81.5  & 79.1 & 76.7 \\
    MORAN-v2 + AEG(ours) & \textbf{94.7} & \textbf{89.6} & $\textbf{82.8}^{\color{red}\uparrow1.3}$ & $81.3^{\color{red}\uparrow2.2}$  & $\textbf{77.4}^{\color{red}\uparrow0.7}$ \\
    \bottomrule
    \end{tabular}%
\end{table}%

\begin{table*}[t]
\caption{Performance comparison on regular benchmarks. ``$50$" and ``$1$k" are lexicon sizes. ``Full" indicates the combined lexicon of all images in the benchmarks. ``None" means lexicon-free. ``*" represents scene text recognition methods with rectification. ``$^\dagger$" indicates the scene text recognition methods with extra datasets.}
\label{Table_regular}
\centering
\scriptsize
    \begin{tabular}{c|ccc|cc|ccc|c}
    \toprule
    \multirow{2}[4]{*}{\textbf{Method}} & \multicolumn{3}{c|}{\textbf{IIIT5K}} & \multicolumn{2}{c|}{\textbf{SVT}} & \multicolumn{3}{c|}{\textbf{IC03}} & \textbf{IC13} \\
\cmidrule{2-10}          & \textbf{50} & \textbf{1k} & \textbf{None} & \textbf{50} & \textbf{None} & \textbf{50} & \textbf{Full} & \textbf{None} & \textbf{None} \\
    \midrule
    Almaz\'{a}n et al. \cite{almazan2014word} & 88.6  & 75.6  & -     & 87.0  & -     & -     & -     & -     & - \\
    Yao et al. \cite{yao2014strokelets} & 80.2  & 69.3  & -     & 75.9  & -     & 88.5  & 80.3  & -     & - \\
    R.-Serrano et al. \cite{risnumawan2014robust} & 76.1  & 57.4  & -     & 70.0  & -     & -     & -     & -     & - \\
    Jaderberg et al. \cite{jaderberg2014deep} & -     & -     & -     & 86.1  & -     & 96.2  & 91.5  & -     & - \\
    Su and Lu et al. \cite{su2014accurate} & -     & -     & -     & 83.0  & -     & 92.0    & 82.0  & -     & - \\
    Gordo et al. \cite{gordo2015supervised} & 93.3  & 86.6  & -     & 91.8  & -     & -     & -     & -     & - \\
    Jaderberg et al. \cite{jaderberg2016reading} & 97.1  & 92.7  & -     & 95.4  & 80.7  & 98.7  & {\textbf{98.6}} & 93.1  & 90.8 \\
    Jaderberg et al.\cite{jaderberg2015deep} & 95.5  & 89.6  & -     & 93.2  & 71.7  & 97.8  & 97.0  & 89.6  & 81.8 \\
    Shi, Bai, and Yao \cite{shi2017end} & 97.8  & 95.0  & 81.2  & 97.5  & 82.7  & 98.7  & 98.0  & 91.9  & 89.6 \\
    *Shi et al. \cite{shi2016robust} & 96.2  & 93.8  & 81.9  & 95.5  & 81.9  & 98.3  & 96.2  & 90.1  & 88.6 \\
    Lee and Osindero \cite{lee2016recursive} & 96.8  & 94.4  & 78.4  & 96.3  & 80.7  & 97.9  & 97.0  & 88.7  & 90.0 \\
    *Liu et al. \cite{liu2016star}  & 97.7  & 94.5  & 83.3  & 95.5  & 83.6  & 96.9  & 95.3  & 89.9  & 89.1 \\
    $^\dagger$Yang et al. \cite{yang2017learning} & 97.8  & 96.1  & -     & 95.2  & -     & 97.7  & -     & -     & - \\
    Yin et al. \cite{yin2017scene} & 98.7  & 96.1  & 78.2  & 95.1  & 72.5  & 97.6  & 96.5  & 81.1  & 81.4 \\
    $^\dagger$Cheng et al. \cite{cheng2017focusing} & 99.3  & 97.5  & 87.4  & 97.1  & 85.9  & \textbf{99.2} & 97.3  & 94.2  & 93.3 \\
    Cheng et al. \cite{cheng2018aon} & \textbf{99.6} & 98.1  & 87.0  & 96.0  & 82.8  & 98.5  & 97.1  & 91.5  & - \\
    *Liu et al. \cite{liu2018char} & -     & -     & 83.6  & -     & 84.4  & \textbf{-} & 93.3  & 91.5  & 90.8 \\
    $^\dagger$Liu et al. \cite{liu2018squeezedtext} & 97.0  & 94.1  & 87.0  & 95.2  & -     & 98.8  & 97.9  & 93.1  & 92.9 \\
    $^\dagger$Bai et al. \cite{bai2018edit} & 99.5  & 97.9  & 88.3  & 96.6  & 87.5  & 98.7  & 97.9  & 94.6  & 94.4 \\
    Liu et al. \cite{liu2018synthetically} & 97.3  & 96.1  & 89.4  & 96.8  & 87.1  & 98.1  & 97.5  & 94.7  & 94.0 \\
    Gao et al. \cite{gao2018dense} & 99.1  & 97.2  & 83.6  & \textbf{97.7}  & 83.9  & 98.6  & 96.6  & 91.4  & 89.5 \\
    *Zhan et al. \cite{zhan2019esir} & \textbf{99.6} & \textbf{98.8} & 93.3  & 97.4  & 90.2  & -     & -     & -     & 91.3 \\
    Zhang et al. \cite{zhang2019sequence} & \textbf{-} & \textbf{-} & 83.8  & -     & 84.5  & -     & -     & 92.1  & 91.8 \\
    *Luo et al. \cite{cluo2019moran} & 97.9  & 96.2  & 91.2  & 96.6  & 88.3  & 98.7  & 97.8  & 95.0  & 92.4 \\
    \midrule
    Baseline  & 99.0  & 97.9  & 92.6  & 96.2  & 87.5  & 98.2  & 97.3  & 93.6  & 92.2 \\
    Baseline + AEG(ours) & 99.4  & 98.3  & $93.6^{\color{red}\uparrow1.0}$  & 96.9  & $89.2^{\color{red}\uparrow1.7}$  & 98.8  & 98.0  & $94.8^{\color{red}\uparrow1.2}$  & $92.9^{\color{red}\uparrow0.7}$\\
    \midrule
    *ASTER \cite{shi2018aster} & \textbf{99.6} & \textbf{98.8} & 93.4  & 97.4 & 89.5  & 98.8  & 98.0  & 94.5  & 91.8 \\
    ASTER + AEG (ours) & 99.5  & 98.5  & $94.4^{\color{red}\uparrow1.0}$   & 97.4  & $90.3^{\color{red}\uparrow0.8}$  & 99.0  & 98.3  & $95.2^{\color{red}\uparrow0.7}$  & $95.0^{\color{red}\uparrow3.2}$ \\
    \midrule
    *MORAN-v2\tablefootnote{Reported results are based on \url{https://github.com/Canjie-Luo/MORAN_v2}}\cite{cluo2019moran} & 99.1     & 98.0     & 94.2  & 97.3     & 89.0  & 98.7     & 98.2     & 95.0  & 95.1 \\
    MORAN-v2 + AEG(ours) & 99.5  & 98.7  & $\textbf{94.6}^{\color{red}\uparrow0.4}$  & 97.4  & $\textbf{90.4}^{\color{red}\uparrow1.4}$ & 98.8  & 98.3  & $\textbf{95.3}^{\color{red}\uparrow0.3}$ & $\textbf{95.3}^{\color{red}\uparrow0.2}$ \\
    \bottomrule
    \end{tabular}%
\end{table*}%

\begin{figure*}[t]
\centering
\includegraphics[width=0.95\textwidth]{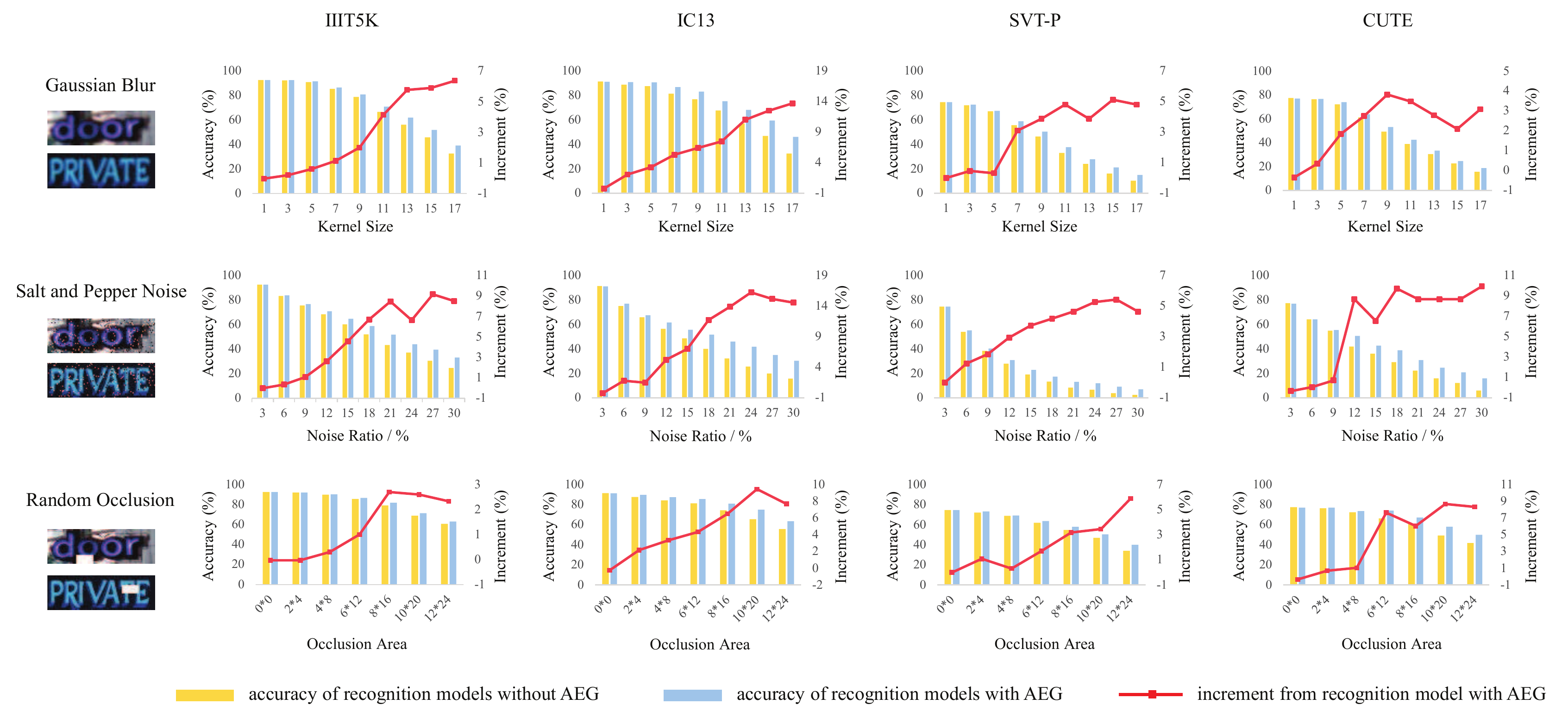}
\caption{Visualization of the text prediction on IIIT$5$k, IC$13$, SVT-P and CUTE datasets under random noise interference.
The red line represents the increment from the recognition models with AEG.}
\label{Figure_robust}
\end{figure*}

\begin{figure}[htbp]
\centering
\includegraphics[width=0.45\textwidth]{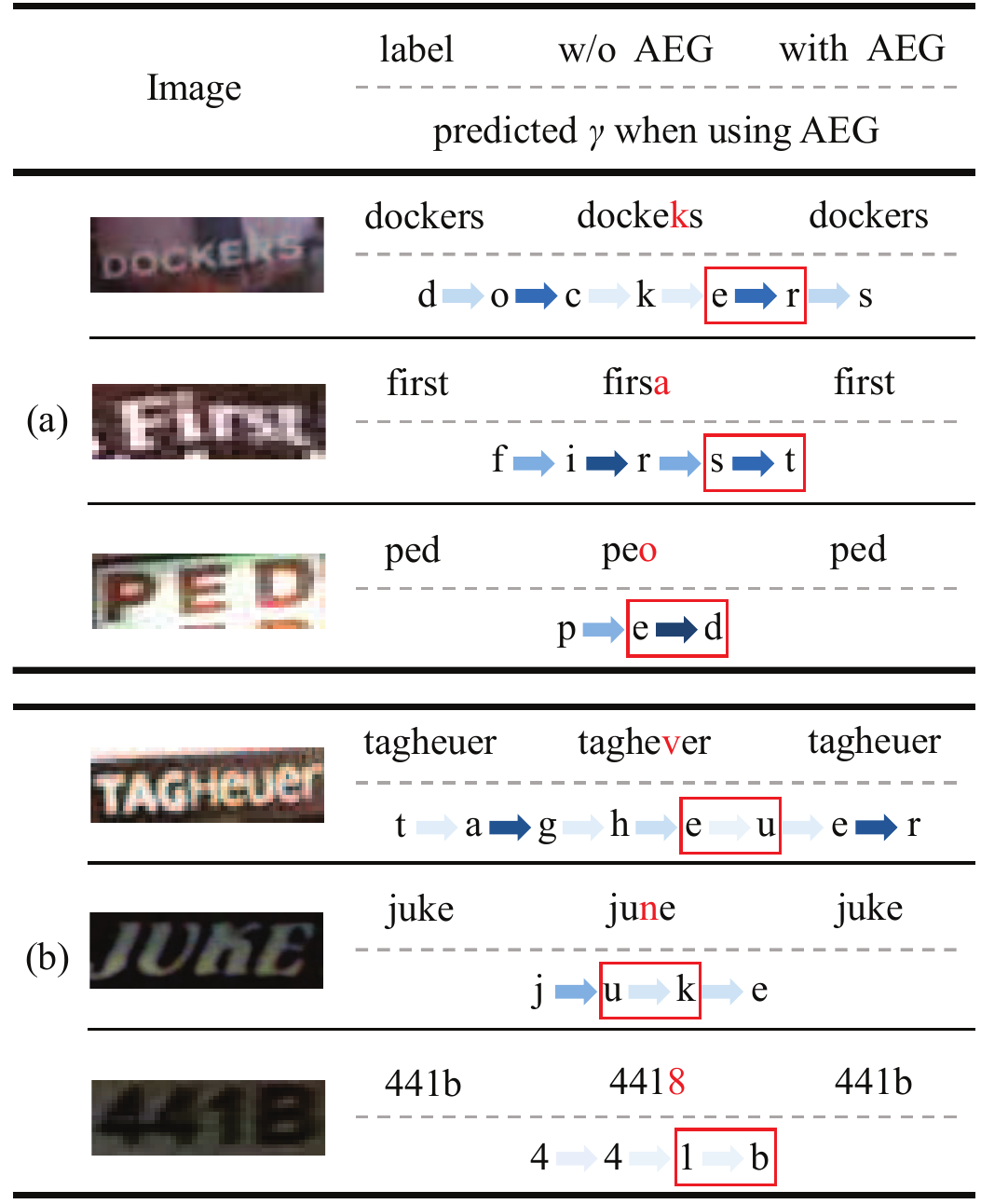}
\caption{
Visualization of some real images recognized without/with the proposed AEG.
The color of arrows indicates the guidance of previous predictions, i.e., the AEG score $\gamma$.
An arrow with deeper color corresponds to the decoding with stronger guidance.
}
\label{Figure_visualization}
\end{figure}

In irregular benchmarks, most of testing samples are low-resolution, perspective and curved text.
The various fonts and distorted patterns of irregular text cause additional challenges in recognition.

As illustrated in Table~\ref{Table_irregular}, AEG significantly improves the performance of baseline on all irregular text datasets.
It should be pointed out that our method does not involve any rectification process which removes the distortions of text and reduces the difficulty of recognition.
However, the baseline with AEG performs competitively on all test benchmarks.

When AEG is integrated to the state-of-the-art frameworks with rectification, recognition performance is also significantly improved.
From Table~\ref{Table_irregular}, it can be seen that by incorporating with AEG, the performance of ASTER \cite{shi2018aster} is improved by $3.5\%$ on SVT-P, $1.4\%$ on CUTE and $0.6\%$ on IC$15$ dataset, respectively.
Similarly, MORAN-v$2$ \cite{cluo2019moran} with AEG outperforms MORAN-v$2$ by $1.3\%$, $2.2\%$, $0.7\%$ on the SVT-P, CUTE, IC$15$ datasets, respectively. 
It indicates that the performance gains brought by the rectification process and the AEG are complementary.

As shown in Table~\ref{Table_regular} and Table~\ref{Table_irregular}, the proposed AEG further improved the performance of state-of-the-arts.
It is clear that our proposed AEG-based model works strongly in general and it demonstrates the performance superiority of AEG.

\subsection{Robustness of AEG}
\label{robust_of_AEG}

Scene text recognition is still challenging owing to the aspects of illumination, low resolution and motion blurring.
However, environmental interference is inevitable.
Thus, performances of text spotting systems in real-world applications are significantly affected by the robustness of text recognition algorithms.
To demonstrate the robustness of AEG, we compare the noise immunity of baseline and AEG-based models on some real images.

IIIT$5$k, IC$13$, SVT-P and CUTE are utilized as the test datasets, and some random noise interferences are added on test images, such as Gaussian blur, salt and pepper noise and random occlusion.
The input distorted images are displayed in the $1$st column in the left part of Figure~\ref{Figure_robust}.

As shown in Table~\ref{Table_regular} and Table~\ref{Table_irregular}, the recognition performance of AEG is significantly better than baseline. 
In order to perform a fair comparison, we limit the performance of AEG by stopping training earlier.
Therefore, the initial performance of obtained AEG model is comparable with that of the baseline model.
As illustrated in Figure~\ref{Figure_robust}, the red line indicates the increment from recognition model with AEG.
As the noise intensity increases, the AEG-based model performs significantly better than baseline on regular and irregular benchmarks.
It indicates the outstanding robustness of AEG under different noise disturbances.

\subsection{Performance Visualization}
\label{performance_visualization}

Figure~\ref{Figure_visualization} shows the results of the performance visualization.
The input images, labels, recognition results of baseline and AEG-based model are displayed in the $1$st, $2$nd, $3$rd and $4$th columns of Figure~\ref{Figure_visualization}, respectively.
The color of the arrow reflects the degree of neighboring character's correlations, i.e., the AEG score.

As illustrated in Figure~\ref{Figure_visualization} (a), AEG-based model can strengthen the correlation of strong-correlated pair and finally correct the recognition results.
Meanwhile, as shown in (b), the AEG can also weaken the guidance weight of previous prediction when it comes the weak-correlated pairs, indicating the adaptiveness of AEG.
From Figure~\ref{Figure_visualization}, it can be seen that the AEG-based model also performs better than baseline on distorted images, such as perspective and blurred text.

\section{Conclusion}
\label{conclusion}

In this paper, we point out the inappropriate use of previous predictions in current existing attention mechanism, which restricts the recognition performance and brings instability.
Hence, we propose an effective and robust module, namely AEG, to introduce proper correlation guidance between adjacent characters by introducing high-order character language model.
In particular, the formulation and three instantiations of AEG are proposed in this paper.
The proposed AEG is flexible and can be easily integrated to existing attention mechanisms.
When incorporating AEG with the state-of-the-art frameworks, AEG can significantly boost their performance.
Extensive experimental results verifies its effectiveness and outstanding robustness under different noise disturbances.
In future, we will apply the AEG idea to other attention-based frameworks in different fields, such as machine translation, speech recognition and image/video caption.

\bibliographystyle{IEEEtran}
\bibliography{refs}

\end{document}